# A Practical Sensing Interface for Exoskeleton Evaluation in Workplaces using Interface Forces *

Joshua Leong Wei Ren, and Thomas M. Kwok, *Member, IEEE*

*Abstract*— This paper presents a novel approach to evaluating back support exoskeletons (BSEs) in workplace settings, addressing the limitations of traditional methods like electromyography (EMG), which are impractical due to their sensitivity to external disturbances and user sweat. Variability in BSE performance among users, often due to joint misalignment and anthropomorphic differences, can lead to discomfort and reduced effectiveness. To overcome these challenges, we propose integrating a compact load cell into the exoskeleton's thigh cuff. This small load cell provides precise force measurements without significantly altering the exoskeleton's kinematics or inertia, enabling real-time assessment of exoskeleton assistance in both laboratory and workplace environments. Experimental validation during load-lifting tasks demonstrated that the load cell effectively captures interface forces between the BSE and human subjects, showing stronger correlations with the user's muscle activity when the BSE provides effective assistance. This innovative sensing interface offers a stable, practical alternative to EMG and respiratory gas measurements, facilitating more accurate and convenient evaluation of BSE performance in real-world industrial and laboratory settings. The proposed method holds promise for enhancing the adoption and effectiveness of BSEs by providing reliable, real-time feedback on their assistance capabilities.

## I. INTRODUCTION

Lower back pain (LBP) is the leading cause of disability worldwide, with a lifetime prevalence rates of between 75% and 84% [1]. Work-related musculoskeletal disorders (MSDs) have resulted in over $200 billion in economic losses due to medical expenses and indirect costs from worker recovery times [2], representing a significant societal burden. High load lifting increases torque on the lumbar region (L5/S1), leading to higher muscle load and tension, which can cause muscle fatigue and subsequent injury [3]. Therefore, reducing the load on lumbar muscles is crucial for lowering the risk of injury.

Today, many back support exoskeletons (BSEs) have been evaluated for different industrial and occupational uses [4, 5], such as airport luggage handling [6-8], automotive industry [9], farm work [10], construction [11], and logistics [12]. These devices provide assistive torque to the trunk and thigh, about the hip joint [13], reducing muscle effort and risk of injury. However, the performance of these BSEs can vary significantly among individuals [8, 14], possibly due to joint misalignment issues. As noted in [15], differences in exoskeleton fit across users may affect the level of assistance provided. Anthropomorphic discrepancies between the user and the exoskeleton may convert part of the assistive force into undesirable parasitic forces [16], leading to discomfort and safety concern.

Moreover, because exoskeleton evaluations are uncommon and not always visible during workplace operations, users may not realize when the exoskeleton is malfunctioning or improperly worn. This underestimation of exoskeleton capabilities can affect their willingness to adopt this assistive technology in the workplace.

Hence, it is essential to evaluate BSE assistance in real-time in workplace settings. Such evaluations can be used not only for functional testing during user experiments [17], but also for quantifying the quality of exoskeleton assistance. Users can know the assistance quality and realize the potential issues about the exoskeleton function or fitting. This allows users to understand the assistance quality and identify potential issues related to exoskeleton function or fit. However, there are limited standards and objective performance metrics for evaluating exoskeleton assistance. The most common evaluation metric, electromyography (EMG) [18], is impractical for field tests or workplace use. As EMG is susceptible to external disturbances, sensor placement issues, and user sweat, which are difficult to avoid in workplace scenarios involving load-lifting tasks.

As suggested in [19], Erezuma et al. highlighted the potential use of interface forces between the user and exoskeleton as an evaluation metric. In existing literature [20-23], most exoskeletons used load cells, optical systems, force sensitive resistors (FSR), and air-based pressure sensors to measure the interface force. But their purposes mainly focus on robot control purposes or safety evaluations such as pain pressure threshold (PPT). To the best of our knowledge, there has been limited evaluation on the quality of assistance provided by the exoskeleton regarding interface forces.

One of direct approaches to interface force measurement is FSRs. For example, some papers used FSRs to investigate assistive forces applied by exoskeletons onto the human body during operation [22, 23]. This was mainly due to the small and flexible nature of FSRs, which allowed them to be integrated in between the human body and rigid exoskeleton interfaces. However, there are trade-offs for these FSRs in the accuracy and precision in force measurements, as well their high hysteresis and poor repeatability due to the drifting of the FSR.

Sposito et al., investigated the parasitic forces at the cuff brace of an exoskeleton by utilizing a Force-Torque (F/T) sensor attached onto the frame of the exoskeleton [24]. However, the F/T sensor used had a large size and weight, which they highlighted that would have altered the original kinematics of the original exoskeleton joint and added additional inertia to the exoskeleton.

This paper proposes a practical sensing cuff for BSEs integrating a small form factor load cell, such as the FX29, for evaluating assistance in workplace settings. This approach balances size, weight, and precision in force measurement, allowing the load cell to be embedded in the exoskeleton's thigh cuff without impacting the kinematics or inertia of the joints. With this compact load cell, it becomes practical to assess exoskeleton assistance in both laboratory and workplace environments.

The rest of the paper is organized as follows. Section II details the proposed sensing interface for interface force measurement. Section III presents the experimental validation conducted during a load-lifting task with two healthy subjects. Finally, Section IV offers a discussion and outlines potential research contributions.

## II. Novel Sensing Interface for Interface Force

### A. Load Cell Specifications

As for sensors for interface force, we utilized a compact load cell (FX29K0 – 100A – 0100 – L (FX29), TE Connectivity Ltd.), which has a maximum load range up to 100 lbf (444.82 N), dimensions with 19.70 mm (diameter) x 5.45 mm (height) and a weight of 6.0 grams.

### B. Sensing Cuff Design

As shown in the Fig. 1, the thigh cuff used in this paper consists of three main components: carbon fibre rod, rod interface attachment and cuff interface. The rod interface attachment is anchored to the cuff interface with 4 recessed nuts in the rod interface and screws through the cuff interface.

As for loadcell implementation, we modified the cuff interface by adding a flat loading panel (80 mm x 50 mm). Such loading puck design takes inspiration from Ghognasi et al. loading panel design for force sensitive resistor (FSR) interface force detection [25]. The rod attachment was also modified to allow for the embedding of the FX29 load cell. In terms of material, the loading puck and modified rod attachment were 3D printed with Black polylactic (PLA) and 40% infill density.

The elongated neck of the loading puck design transmits force to the load cell from an interface plate that contacts the human body. The elongated neck allows the load interface to pass through the hole in the center of the cuff. The top of the elongated neck has a narrower portion to allow for small tilting of the loading puck, ensuring the effective force loading to FX29 load cell. The height of the elongated loading puck neck was empirically tuned to reduce the tilting of the combined structure.

Adhesive neoprene cushion material and a thin silicone dampener was applied between the load applicator and the inner face of the interface brace to prevent rotational movement of the load applicator as well as to function as a soft cushioning material which would not interfere with the compressive forces applied on to the load applicator by the exoskeleton.

As shown in Fig. 1, the rod interface for the cuff was modified to have a recessed portion (6 mm) for the FX29 load cell to be embedded into the centre of the rod interface. This recess aligned over the opening of the cuff brace and was orientated in such that the cable can be fed through the carbon fibre rod to remove any form of interference with the cuff as it rotates (Fig. 2).

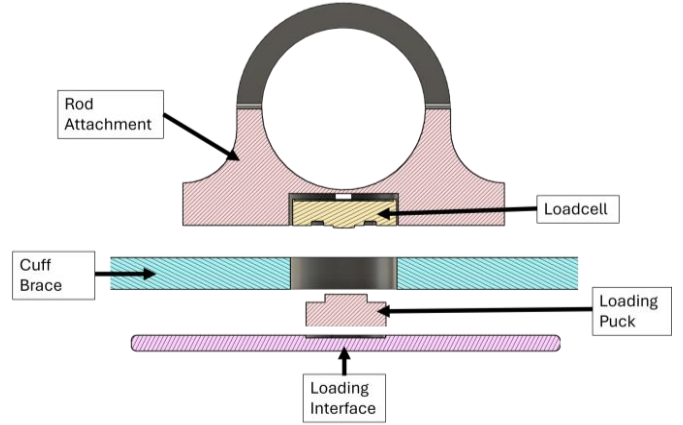

Fig. 1. Cross section of cuff sensing solution

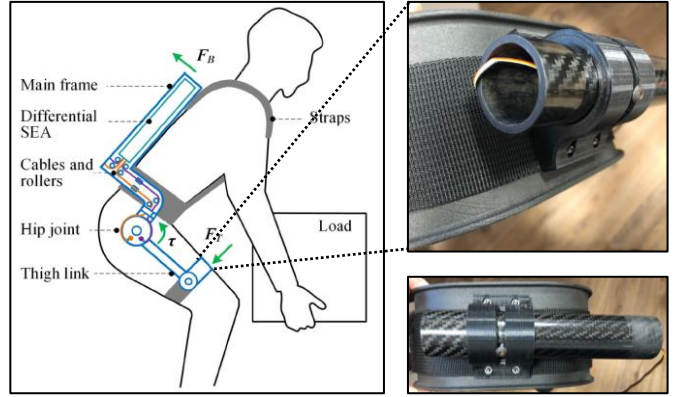

Fig. 2. Integrated sensing cuff showing cable routing and working principle of BSE

### B. Sensing System Design

FX29 loadcell output cables were soldered onto SparkFun Qwiic Cables with 1mm JST termination, in accordance with I$^2$C connections. A SparkFun Qwiic adapter was soldered with 20pF ceramic capacitors across the SCL and GND plated through holes on the adapter, this was to increase the I$^2$C communication clock signal resistance to noise due to the increased cable length to reach the electronic housing body on the exoskeleton and the proximity of the I$^2$C cable run to the exoskeleton motor. A multiplexor (SparkFun Qwiic Mux Breakout – 8 channels, TCA9548A) was used to read the left and right load cell outputs through the single I$^2$C channel of the IMU board. And the IMU data was collected by IMU board (STM32F446xC IC).

The example electrical connections can be seen in Fig. 3 for a single load cell configuration for clarity. Subsequently, FX29 loadcell data from multiplexor and IMU data were sent to the IMU board through I$^2$C communication protocol. After that, loadcell and IMU data were sent to the main processor board (STM32F446xC IC) through Controller Area Network (CAN bus) communication protocol. The main processor board contains the control algorithm and processing code for the whole BSE system. As for data collection, the main exoskeleton processor board transferred the loadcell data to PC via serial port connection. The main exoskeleton processor queries for data across the CAN bus at 500 Hz, with the IMU and FX29 data matched accordingly.

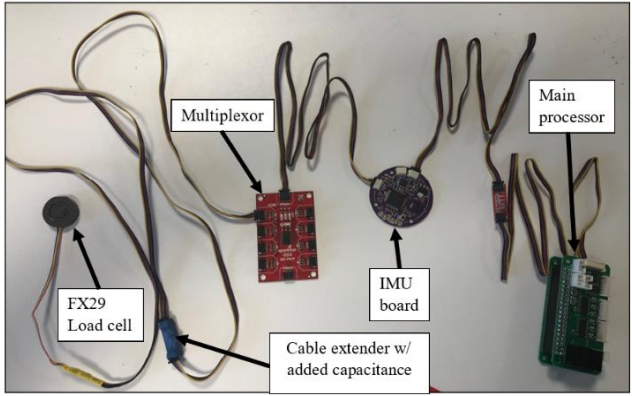

Fig. 3. The electrical connections for a single load cell configuration.

## III. Experiment Validation

### A. Sensor Validation

Load cells utilize strain gauges to convert the load acting onto them into small electrical signals and would require pre-amplification. The relationship between the digital output and the applied force is defined by the following Equation (1):

$$F = (O - Z) * \frac{R}{14000} \quad (1)$$

where $F$ is the loading force output in pounds-force (lbf), $O$ is digital decimal output of the FX29 loadcell, $Z$ is digital decimal zero offset output of the FX29 loadcell, and $R$ is its maximum load range (100 lbf). The force output, $F$ was then converted from pounds-force into newtons (1 lbs = 4.44822 N).

For each verification trial, six known weights of different masses were placed over the sensor. This was done for a total of 5 trials and each trial was sampled for a duration of 10 seconds. A force distribution puck was positioned between the FX29 sensor and the applied weight to allow even distribution of the weight over the sensor (Fig. 4). The different sensor combination tested was FX29 load cell alone, and FX29 load cell with a cushion placed on top the force distribution puck. The weights used were 1kg, 2kg, 3kg, 4kg, 5kg and 6kg weight plates.

From the mean testing, the data has been tabulated in Table 1. The small value of standard deviation suggests that the measurement of data is repeatable and robust to small distributions in the load distribution. Errors for the loading interface with cushioning were all under 0.5 N and the largest deviation was – 2.24% for the 1 kg load. Furthermore, the force error of the load cell with the cushioning was not greater than without the cushioning.

Table 1. Cushioning Effect on Load Cell Calibration

| Load (kg) | without cushioning | | with cushioning | |
|---|---|---|---|---|
| | Force (N) | Error (%) | Force (N) | Error (%) |
| 1 | 9.25 ± 0.030 | -5.75 | 9.59 ± 0.005 | -2.24 |
| 2 | 19.48 ± 0.007 | -0.70 | 19.65 ± 0.005 | 0.15 |
| 3 | 29.53 ± 0.009 | 0.33 | 29.64 ± 0.003 | 0.73 |
| 4 | 38.82 ± 0.012 | -1.06 | 39.04 ± 0.012 | -0.51 |
| 5 | 48.57 ± 0.003 | -0.98 | 48.81 ± 0.006 | -0.48 |
| 6 | 57.79 ± 0.030 | -1.82 | 58.32 ± 0.014 | -0.93 |

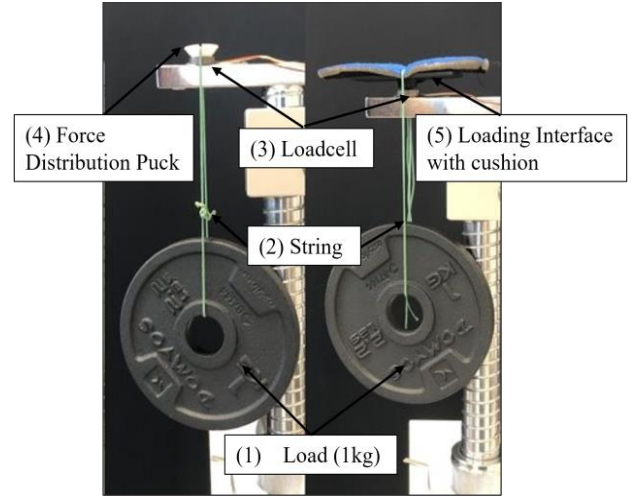

Fig. 4. Load cell calibration setup with and without cushion. (1) Load, (2) string, (3) loadcell, (4) force distribution puck (5) loading interface with cushion.

### B. Exoskeleton Implementation

The details of the BSE used in this study can be found in our previous work [26]. The robot assistance is provided by a differential series elastic actuator (D-SEA) linked to the hip joint and thigh cuff. This BSE generates an assistive torque for hip joints and the load is distributed into a supportive force on the shoulders ($F_B$) and a compressive force onto the user's thighs ($F_T$) as shown in Fig. 2.

For the following human experiment, the dampening coefficient of the exoskeleton was fixed for all subjects, providing them with a similar level of assistance through the experiments.

### C. Experiment Protocol

In the experiments, we recruited 2 healthy male participants (Subject 1, age: 22 years old, height: 169 cm, weight: 56 kg; Subject 2, age 23 years old, height: 170 cm, weight: 70 kg). The inclusion criteria were given that they did not have a history of LBP or upper limb injury at the start of the experiment. The experimental protocol was approved by the Institutional Review Board of the National University of Singapore (NUS-IRB Study LK-20-021). Participants provided written informed consent on the day of the experiment.

In our experiment, we aimed to test the functionality of the sensing cuff and verify it by comparing it with the surface electromyography (EMG) signal and motor torque. Since the BSE is designed to assist user when needed [26], the exoskeleton force should assist the motion required muscle activations, and should not resist the motion that is not required muscle activations. Hence, we expected to observe that the interface force captured by sensing cuff will be correlated with muscle activation that represented by EMG signal. With the correlation between interface force and robot assistance, we can confirm the interface force is expected as our BSE design purpose, and it is not introduced by users' random motion.

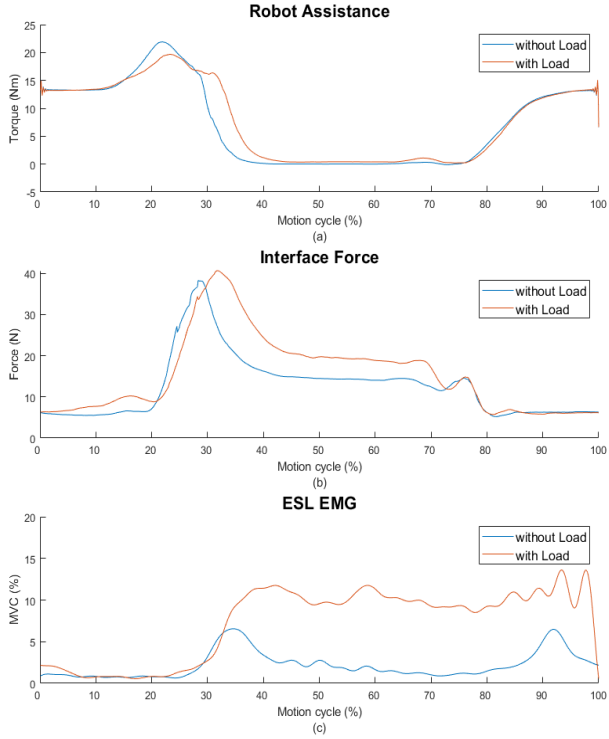

Fig. 5. Comparison between (a) robot assistance, (b) interface force, and (c) ESL muscle EMG signal when subject 1 performs the stoop lifting and lowering motion.

Before the study was conducted, the subjects' basic anthropomorphic data were collected, and adjusted the exoskeleton size accordingly. The subjects were given 10 minutes to familiarize themselves with the exoskeleton movement and assistive force.

The task involved lifting and lowering a 10 kg load placed within a plastic box. The task can be broken down into 2 sequential motions: lifting and lowering without load, and lifting and lowering with load, as shown in Figure 5.

The lifting and lowering rate was set at a pace of 6 cycles per minute for a duration of 5 minutes, totaling 30 lifting/lowering cycles per condition. The pace was maintained by the metronome. 10 uninterrupted cycles were taken for analysis from the total duration.

The subjects were instructed to completed with task with stoop lifting (STP), where the knees are kept as straight as possible. The tasks were supervised to ensure that the subjects were maintaining the correct lifting technique. A minimum of 15 minutes of rest time was given to the subjects between each task, with more time given to the subjects if requested.

During the experiment, four EMG (Delsys Trigno Sensor) were recorded bilaterally from two different muscle groups in the lower back regions, the erector spinae longissimus (ESL), and erector spinae iliocostalis (ESI). The sensors have a sampling frequency of 2000 Hz, and an amplifier gain of 1000. The EMG signals were processed as follows: (i) filtered using a fourth-order bandpass Butterworth filter with a 10-400 Hz bandwidth, (ii) detrended and rectified, (iii) smoothed with a moving average over a 0.2-second window, (iv) normalized to each subject's maximum voluntary contraction (MVC), and (v) averaged between each respective left and right muscle groups. An IMU sensor (Delsys Trigno Sensor) was placed onto the weighted box using double sided tape to detect the impact of the box onto the floor as an event detection.

Data from the exoskeleton system was captured at 500,000 Hz baud rate. The extracted data contained the exoskeleton trunk IMU data and the load cell force data from the embedded left and right thigh cuffs. The peak trunk angle of the exoskeleton was used as the epoch event detection for cyclical analysis of the loadcell force data and normalized to the motion cycle. In each lifting motion, the force data from left and right load cells were averaged together to form a singular interface force (IF) profile.

Table 2. Pearson Correlation ($r$) between Interface Force (IF) and Robot Assistance (RA) when Stoop Lifting with (w) and without (wo) Load

| Subject | Load | $r$ | $p$ |
|---|---|---|---|
| 1 | wo | 0.80 | <0.0001 |
| 1 | w | 0.78 | <0.0001 |
| 2 | wo | 0.61 | <0.0001 |
| 2 | w | 0.58 | <0.0001 |

Table 3. Pearson Correlation ($r$) between Interaction force and Muscle Activation when Stoop Lifting with (w) and without (wo) Load

| Subject | Load | Muscle | $r$ | $p$ |
|---|---|---|---|---|
| 1 | wo | ESL | 0.57 | <0.0001 |
| 1 | w | ESL | 0.64 | <0.0001 |
| 1 | wo | ESI | 0.67 | <0.0001 |
| 1 | w | ESI | 0.71 | <0.0001 |
| 2 | wo | ESL | 0.18 | <0.0001 |
| 2 | w | ESL | 0.33 | <0.0001 |
| 2 | wo | ESI | 0.16 | <0.0001 |
| 2 | w | ESI | 0.17 | <0.0001 |

*D. Experiment Result*

Pearson correlations between the IF and robot assistance (RA) had strong positive correlation ($r > 0.5$) and were statistically significant ($p < 0.0001$) for all subjects, with and without load, as shown in Table 2.

Subject 1 Pearson's correlations between the IF and muscle activation all showed a statistically significant strong positive correlation ($r > 0.5$, $p < 0.0001$), regardless of load lifting or back muscle. Subject 2 Pearson's correlation instead had statistically significant moderate positive correlation ($r = 0.33$, $p < 0.0001$) for the ESL muscle versus with load condition. All other load lifting and back muscles had a statistically significant weak positive correlation ($r < 0.2$, $p < 0.0001$).

Fig. 5. shows the comparison between IF, robot assistance (RA), and EMG signal when subject 1 performed the stoop lifting and lowering motion. We can observe a similar tendency of interface force with robot assistance and EMG signal. Since IF is correlated with RA, we confirm that the interface force is as expected as we command BSE, and it is effectively transmitted to subjects. The interface force captured by the sensing cuff is correlated with muscle activation during lifting motion.

## IV. DISCUSSION

The robot assistive torque having a strong statistical correlation with the IF in subject 1, suggests that loadcells can capture the BSE assistance, given that they measured a portion of the forces applied to the user's body. This means that the loadcell system can be a reliable representative of the BSE assistance across subjects. Analyzing the correlation between the IF and EMG, having a strong correlation suggests that the IF has a similar tendency as the EMG profile.

However, having a weak correlation between the IF and EMG in subject 2 could suggest that the assistance from the robot assistance has not been effective due to the generated parasitic forces, which would require some form of adjustment or maintenance of the BSE to rectify. This means that there are situations where, if the BSE is fitted appropriately, the IF measurements can be used as a metric similar to EMG to inform the BSE assistance performance.

Overall, these insights could be promising for industrial BSE evaluations, as having the IF can provide a more stable form of BSE evaluation, as the IF measurements and loadcells would not be as susceptible to damp skin conditions caused by user sweat as compared to EMG sensors. Furthermore, as the loadcells are integrated into the BSE cuff, it would not require extra preparation time to implement and capture the IF measurements. This would be beneficial for collecting performance data in industrial and workplace settings, as EMG sensors would require time to prepare the subject's skin, obtain baseline muscle activations to determine MVC and measures must be put in place to reduce interference from user sweat disrupting the EMG sensor electrode contact, which can be difficult in warm workplace conditions while doing manual handling tasks. Respiratory gases measurement for metabolic cost (MC) evaluation is also a measure for workload intensity [27], and has been utilized in other BSEs evaluations [14, 19, 28, 29], while overcoming some of the hurdles of EMG sensors. However, while the device for respiratory gas measurements have become portable [30], they still require a gas calibration period, a fitted mask on the user, and a minimum trial duration of 5 minutes to achieve a metabolic steady state which would extra time and constraints to collecting the MC data in workplace settings.

Despite the promising results presented in this paper, we cannot yet conclude that interface force (IF) measurements can fully replace EMG for evaluating BSE assistance at this exploratory stage. However, IF shows potential in assessing exoskeleton assistance. A larger subject population is needed to further investigate the correlation between IF and EMG, ensuring that the findings are representative of the broader working population.

These results and the integrated loadcell system could have possible implementations for other rehabilitative exoskeletons as well, whether for upper limb [25, 31, 32] or lower limb [23, 33].

Furthermore, it is interesting to observe in Fig. 5, the load cell force profile showed increased interface force during the standing phase with load compared to without load. The BSE assistance would not be engaged as the upright trunk angle would have placed in transparency mode [26]. This could be an important insight to pursue as this would mean that the BSE with the embedded load cell system would be able to detect the user carrying a load, which in turn may improve the controller design for the applied torque profile of the BSE in future.

## V. CONCLUSION

In this paper, we developed a practical sensing cuff for BSEs integrating a load cell for evaluating assistance in workplace settings. This approach balances size, weight, and precision in force measurement, allowing the loadcell to be embedded in the exoskeleton's thigh cuff without impacting the kinematics or inertia of the joints. With this compact load cell, it becomes practical to assess exoskeleton assistance in both laboratory and workplace environments. The sensing cuff is not affected by other evaluation sensors constraints, such as user sweat or externally attached devices and would require no additional time in setting up for data collection, beyond fitting the exoskeleton onto the user. This is promising in workplaces or where the evaluation environment is not conducive to EMG or respiratory gas measurements.